# Learning Position Evaluation Functions Used in Monte Carlo Softmax Search

HARUKAZU IGARASHI[†1]   YUICHI MORIOKA
KAZUMASA YAMAMOTO

*Abstract*: This paper makes two proposals for Monte Carlo Softmax Search, which is a recently proposed method that is classified as a selective search like the Monte Carlo Tree Search. The first proposal separately defines the node-selection and backup policies to allow researchers to freely design a node-selection policy based on their searching strategies and confirms the principal variation produced by the Monte Carlo Softmax Search to that produced by a minimax search. The second proposal modifies commonly used learning methods for positional evaluation functions. In our new proposals, evaluation functions are learned by Monte Carlo sampling, which is performed with the backup policy in the search tree produced by Monte Carlo Softmax Search. The learning methods under consideration include supervised learning, reinforcement learning, regression learning, and search bootstrapping. Our sampling-based learning not only uses current positions and principal variations but also the internal nodes and important variations of a search tree. This step reduces the number of games necessary for learning. New learning rules are derived for sampling-based learning based on the Monte Carlo Softmax Search and combinations of the modified learning methods are also proposed in this paper.

## 1. Introduction

Recently, *Monte Carlo Tree Search* (MCTS) [1], which is a type of selective search, has been applied to computer *shogi* (Japanese chess). Based on its wide success, it has become one of the leading methods applied to computer *go*, and Alpha Zero, which is based on MCTS, scored fairly well against the strongest programs based on traditional search algorithms even in chess and shogi [2].

MCTS's typical search method is the UCT algorithm (Upper Confidence Bound 1 Applied to Trees) [3], which deterministically selects a child node with the largest upper confidence bound (UCB) among its brother nodes. The *Monte Carlo Softmax Search* (MCSS) is another proposed selective search scheme [4]. Continuous functions like Boltzmann distribution functions are used for node selection and backup calculation in MCSS, unlike for MCTS. That makes it possible to calculate the gradient vectors of the positional evaluation functions with respect to the parameters included in those functions to introduce a new dimension that is not expected from MCTS.

This paper makes two proposals on MCSS, (1) separation between a node-selection policy and a backup policy, and (2) new sampling-based learning algorithms performed in a search tree produced by MCSS. In MCSS, a policy for selecting a child node while searching used the same Boltzmann distribution function where a backup policy propagates the values of the leaf nodes toward the upper nodes [4]. This step caused a problem: the optimal sequence of the moves estimated by MCSS with a finite temperature parameter did not always conform to the *principal variation* (PV) defined in the minimax search. We separately define the two policies and use a backup policy with a low temperature to find the PV. A node-selection policy is freely designed to control the depth and width of a search tree based on the system designer's strategies.

Second, we show that positional evaluation functions can be learned by Monte Carlo sampling with backup policies in a search tree produced by MCSS. Many methods have been proposed for the learning evaluation functions of computer chess and shogi. Most works are based on supervised learning, reinforcement learning, regression learning, and search bootstrapping. These learning methods (except bootstrapping) only exploit the current positions and the *principal leaves* after the legal moves at the current positions. Current position refers to a state that actually appears in games for learning, and principal leaf denotes a leaf node of a PV. This paper pays attention not only to PV but also to the leaves of the important variations around it, which means that the feature parameters, which are highly related to the positions of those leaf nodes, can be updated toward their optimal values. Internal nodes in a search tree are learned by bootstrapping.

Usually for learning positional evaluation functions, the parameters to be learned must be included in or highly related to the positions prepared for learning. A huge amount of learning data, such as the records of games between professional shogi

---

[†1] Shibaura Institute of Technology

players or the self-play games of the learning system, must be prepared to increase the learning accuracy. However, the learning method proposed in this paper will considerably reduce the number of games required for learning.

For explanatory simplicity, we mainly restrict the scope of our theory's application to computer shogi in this paper.

## 2. General Flow of Monte Carlo Softmax Search

In the search method of MCSS [4], a softmax operation selects a child node based on a certain probability distribution function instead of a deterministic minimax calculation over the values of the child nodes. The serial processes that stochastically select the child nodes from the root node down to a leaf node in a search tree are considered as Monte Carlo sampling. For this reason, the method proposed in Ref. 4 is called a Monte Carlo Softmax Search (MCSS).

The following is the general MCSS flow [4][a]:

1) Initialization: Set current state $u$ to the root node.
2) Selection: One child node $v$=child ($u$) is selected by a stochastic node-selection policy.
3) Expansion: If $v$ has already been expanded, then go to 2) after replacing $u$ with $v$. Otherwise, expand all the child nodes of $v$ of only one step and evaluate them by a quiescence search.
4) Backup: Trace back the path from $v$ to the root node and update all the node values on the path.

MCTS has almost the same flow. However, it uses a deterministic node-selection policy in step 2 and uses *playout* for evaluating the leaf nodes in step 3. The detailed processes and characteristics of MCSS are described in the next section.

## 3. Details of MC Softmax Algorithm

### 3.1 Softmax operation in a backup policy

In shogi, the search's objective is to determine the principal variation. The deeper the search, the more accurate a PV is found under the constraints of the thinking time. An iterative deepening search algorithm is usually used to expand all the next child nodes in only one step. In this algorithm, the child node values are back-propagated to calculate the values of their parent nodes. The leaf nodes of the search tree are evaluated by a positional evaluation function. The updating procedure of the values of the internal nodes is called *backup* and is repeated up to the root node.

Minimax calculation is used in the backup process to search for the PV. Instead of minimax calculation, the MC softmax search uses the following softmax operation that defines an internal node's value using the expectation of its child node's values as follows:

$$E_s(s) = \sum_{a \in A(s)} P(a;s)\, E_s(v(a;s)), \quad (3.1)$$

where $E_s(s)$ is the value of node $s$, $A(s)$ is a set of legal moves at $s$, and $v(a;s)$ is a child node produced from $s$ by move $a \in A(s)$. Weight $P(a;s)(\geq 0)$ is a probability distribution function that satisfies

$$\sum_{x \in A(s)} P(x;s) = 1. \quad (3.2)$$

In steps 1 to 4 of the search algorithm in Section 2 [4], node-selection policy function $\pi(a|s)$ was defined by a Boltzmann distribution function with its children's values, $E_s(v(a;s))$. The same function was used as $P(a;s)$ in Eq. (3.2). In this paper, we call $P(a;s)$ a *backup policy* to distinguish it from node-selection policy $\pi(a|s)$ in step 2 [b]. $P(a;s)$ gives the weight values in the expectation operator that propagates the values of the child nodes up to their parent nodes in step 4.

Minimax calculation only propagates the value of the minimum/maximum node to its parent node. The information on the important variations is completely discarded other than a PV. Moreover, minimax calculation complicates propagating the values of the gradient vectors toward the root node. Even a small variance of the parameters included in an evaluation function discontinuously changes a PV's path and the derivatives of the root node's value. On the other hand, a softmax operator makes it possible to express the values of all the nodes as the continuous functions of the parameters in an evaluation function. This is a desirable property as a backup operator because calculating the gradient vectors of the node values is necessary for learning a positional evaluation function.

Therefore, the softmax operator in a backup policy makes it possible to consider the leaf nodes of important variations other than a PV when learning an evaluation function [c]. In Sections 4 and 5, we show that Monte Carlo sampling with the backup policy on a search tree produced by MCSS can take account of the important variations in supervised learning, reinforcement learning, regression learning, and search bootstrapping.

### 3.2 Stochastic selective search in node selection

Minimax search, which intends to find the PV in a whole

---

a) The *Shibaura Shogi Softmax* team used MCSS in The 27th World Computer Shogi Championship in 2017.
b) The term, *Softmax Policy,* is used in reinforcement learning as a policy through which an agent selects its action. The node-selection policy corresponds to an agent's action selection policy.
c) We can use a minimax operator or set $T$ at a small value close to zero when no learning is being done.

search tree defined by minimax calculation, is classified as a *full-width search*. This approach is one of the leading search algorithms for shogi. A search that restricts the searching region by heuristics is called a *selective search*. *Gekisashi* is a typical shogi program that uses selective search based on the heuristic features of moves [5].

In computer *go*, MCTS is the most successful searching method [1]. UCT is a typical example of MCTS and deterministically selects a child node with the largest upper confidence bound (UCB) among its brother nodes [3]. UCT is a deterministic selective search method. AlphaZero [2] proved that MCTS is applicable even to chess and shogi as well as *go* [d].

Recently selective search is being applied to shogi more extensively. Many researchers are interested in whether such a new approach as MCTS outperforms traditional approaches based on full-width searches like an alpha-beta search. MCSS (explained above in Section 2) is classified in the selective search as MCTS. However, MCSS is different from MCTS because it uses positional evaluation functions to calculate the values of the leaf nodes instead of the playout simulations conducted in MCTS. A more different point is that MCSS stochastically selects a child node with node selection probability $\pi(a|s)$ [e]. In UCT, which is a typical MCTS method, the upper bounds of the child nodes are estimated [3]. Upper bound $I_a$ (UCB1) [7] is defined by

$$I_a \equiv \bar{X}_a + c\sqrt{\frac{2\ln n_s}{n_a}}, \qquad (3.3)$$

where $n_s$ and $n_a$ are the number of visiting $s$ and selecting $a$ at $s$. $\bar{X}_a$ is the sampling average of the payoff after selecting $a$ up to this time [7][8], and $c$ is a weight coefficient that provides a balance between exploitation and exploration [f].

The child node with the largest $I_a$ value among its brother nodes is selected deterministically [3][g]. Upper bound $I_a$ includes the number of visiting nodes so that *infrequently* visited nodes are preferentially selected. However, the MCSS algorithm directly uses $\bar{X}_a$ instead of estimated upper bound $I_a$ and selects a child node in a stochastic manner. Temperature parameter $T$ and strategies with heuristics on the game can be embedded into the node selection policies. That controls the depth of the search and the balance between exploitation and exploration and expresses playing styles [h].

The stochastic selective search in MCSS has two advantages. First, it is suitable for parallel and autonomous computing. A series of selecting nodes from the root node to a leaf node is a stochastic sampling process. These sampling calculations can be executed asynchronously in parallel. Stochastic selection prevents the same node being accessed at a time. It also has an advantage in mini-batch searching. Parallel searching in MCSS is a multiagent system where search agents share the search tree as a common memory like a blackboard. The search tree can be completely divided and assigned to agents.

Second, explicit pruning techniques in a search tree, which are devised in the full-width searches as an alpha-beta search, are not necessary in MCSS. Nodes with low scores are not selected so frequently, and the production of arcs from the nodes is naturally suppressed.

**3.3 Boltzmann distribution functions applied to policies**

In thermodynamics and statistical physics, the principle of maximum entropy proposed by Jaynes derives the probability distribution function necessary for calculating such physical quantities as the energy of a physical system [9]. In this section, we show that the principle also derives a Boltzmann distribution function and can be applied to the stochastic action selection of agents.

Consider the problem of how an agent selects an optimal action at the current state using heuristics $E$ that evaluate the agent's actions. $E_i$ refers to the score of action $i$, which is selected with probability $p_i$ calculated from $E_i$. The value of expectation $<E>$ of $\{E_i\}$ (i=1,2,…,n) is specified to value $E_T$ to maintain the selection quality.

The principle of maximum entropy argues that an action should be selected based on probability function $p_i$ that maximizes entropy $f(\{p_i\}) \equiv -\sum_i p_i \ln p_i$ under the constraints imposed on $\{p_i\}$ and maintains the largest uncertainty about action decisions. However, that decision seems the most unbiased. The Lagrange multiplier method gives the Boltzmann distribution function:

$$p_i = e^{E_i/T} \bigg/ \sum_{i=1}^{n} e^{E_i/T} \qquad (3.4)$$

by maximizing entropy $f(\{p_i\})$ under constraints $\sum_i p_i = 1$ and $\langle E \rangle \equiv \sum_i p_i E_i = E_T$ [4][i].

When temperature $T$ approaches zero, only $p_i$ of action $i$ that has the largest $E_i$ increases to 1 and the stochastic selection policy in (3.4) is reduced to a deterministic policy [j]. On the other hand, the policy approaches a random policy at infinite temperature.

---

d) Alpha Zero used a neural network model for approximating the positional evaluation functions. This model does not go well with sequential processing as an alpha-beta search because it takes too much computing time.
e) Even in MCSS, playout can be used for node evaluation, and learning the simulation policy used in the playout is proposed in Ref. 6.
f) c is set to 1 in Ref. 7 that proposed the UCB1 algorithm.
g) As the visiting number increases, sample average $\bar{X}_a$ approaches the true value and the second term in Eq. (3.3) gradually reduces.
h) A node-selection policy can be defined independently from a backup policy, as will be mentioned in Section 3.3.
i) $T$ is calculated by Eq. (3.4) and $\sum_i p_i E_i = E_T$. Therefore, $T$ or $E_T$ is a measure of the randomness of exploration.
j) The policy becomes a deterministic selection to minimize $E_i$ if $E_i$ is replaced with -$E_i$.

The stochastic policy for selecting an action based on the Boltzmann distribution function is widely used as *Boltzmann selection* in reinforcement learning and as the *Boltzmann-exploration-based search* in multiarmed bandit problems [3][10].

In this paper, we propose two Boltzmann-type functions for node-selection policy $\pi(a|s)$ and backup policy $P(a_i;s)$. However, $E_i$ and $T$ are not necessarily identical in the two functions.

### 3.4 Recursive definition of move and node values

Let $a_t$ be a move at the *t*-th turn ($t=1,2,\ldots,L_a$) in a two-player game like shogi. In MCSS, backup policy $P_a(a_t;u_t)$ is given by a Boltzmann distribution function:

$$P_a(a_t;u_t) = exp(Q_a(u_t,a_t)/T_a)/Z_a \tag{3.5}$$

$$Z_a \equiv \sum_{x \in A(u_t)} exp(Q_a(u_t,x)/T_a). \tag{3.6}$$

$T_a$ is a temperature parameter. $Q_a(u,a)$ is called a *move value*, which indicates the value of move *a* at position *u*.

Suppose that the opponent's backup policy is given by $P_b(b_t;v_t)$, where $b_t$ is the opponent's move at the *t*-th turn (Fig.1). $d$ (=0,1,$\cdots$,D) means the searching depth in Fig. 1. Superscripts are omitted when $d$=0. For example, $u_t^0$ is denoted as $u_t$. $P_b(b_t;v_t)$ is given by

$$P_b(b;v) = exp(-Q_b(v,b)/T_b)/Z_b \tag{3.7}$$

$$Z_b \equiv \sum_{x \in A(b)} exp(-Q_b(v,x)/T_b) \tag{3.8}$$

using move value function $Q_b(v,b)$ [k].

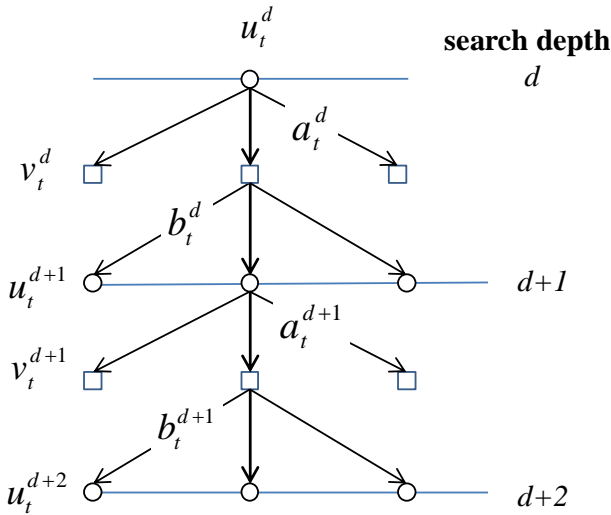

Fig.1 Search tree

---

k) The backup policy and the move value function of the opponent agents are not necessarily required to satisfy Eq. (3.7).
l) Transition probabilities are necessary if the state transition is probabilistic after making a move.

We define *node values* $V_a(u_t^d)$ and $V_b(v_t^d)$ in the search tree using move values:

$$V_a(u_t^d) \equiv \sum_{a_t^d} P_a(a_t^d;u_t^d)Q_a(u_t^d,a_t^d) \tag{3.9}$$

and

$$V_b(v_t^d) \equiv \sum_{b_t^d} P_b(b_t^d;v_t^d)Q_b(v_t^d,b_t^d). \tag{3.10}$$

Since a state after a move is determined uniquely in shogi, move value $Q_a(u,x)$ is replaced by node value $V_b(v(x;u))$ as

$$Q_a(u_t^d,a_t^d) = V_b(v_t^d) \tag{3.11}$$

and

$$Q_b(v_t^d,b_t^d) = V_a(u_t^{d+1}), \tag{3.12}$$

where $v(x;u)$ is the opponent's state after move *x* at *u* [l]. Using Eqs. (3.9) to (3.12), the move values are written recursively:

$$Q_a(u_t^d,a_t^d) = V_b(v_t^d) \tag{3.13}$$

$$= \sum_{b_t^d} P_b(b_t^d;v_t^d)Q_b(v_t^d,b_t^d) \tag{3.14}$$

$$= \sum_{b_t^d} P_b(b_t^d;v_t^d)V_a(u_t^{d+1}) \tag{3.15}$$

$$= \sum_{b_t^d} P_b(b_t^d;v_t^d) \sum_{a_t^{d+1}} P_a(a_t^{d+1};u_t^{d+1})Q_a(u_t^{d+1},a_t^{d+1}) \tag{3.16}$$

When node $u_t^d$ is a leaf node, i.e., $d=D-1$, the move value is defined using positional evaluation function $H_a(u;\omega)$ [m]:

$$Q_a(u_t^{D-1},a_t^{D-1}) \equiv \sum_{b_t^{D-1}} P_b(b_t^{D-1};v_t^{D-1})H_a(u_t^D). \tag{3.17}$$

Similarly, node values are written recursively from (3.7) to (3.10):

$$V_a(u_t^d) = \sum_{a_t^d} P_a(a_t^d;u_t^d)Q_a(u_t^d,a_t^d)$$

$$= \sum_{a_t^d} P_a(a_t^d;u_t^d)V_b(v_t^d) \tag{3.18}$$

$$= \sum_{a_t^d} P_a(a_t^d;u_t^d) \sum_{b_t^d} P_b(b_t^d;v_t^d)Q_b(v_t^d,b_t^d) \tag{3.19}$$

$$= \sum_{a_t^d} P_a(a_t^d;u_t^d) \sum_{b_t^d} P_b(b_t^d;v_t^d)V_a(u_t^{d+1}). \tag{3.20}$$

If node $u_t^D$ is a leaf,

$$V_a(u_t^D) \equiv H_a(u_t^D;\omega). \tag{3.21}$$

Recursive Eqs. (3.16) and (3.20) calculate the backup operations of the move and node values if backup policy functions $P_a(a;u)$

---

m) Eq. (3.14) is a recursive definition of the node values, and Eq. (3.5) is an example of the backup policy functions based on the node values.

and $P_b(b;v)$ are given. $H_a(u;\omega)$ is assumed to be a linear function of the position's features or a neural network function with weight vector $\omega$.

Stepping down to bottom node $u \in U_D(u_t, a_t)$ of the recursion in Eqs. (3.16) and (3.20), move values $Q_a(u_t, a_t)$ and node values $V_a(u_t)$ are expressed using transition probabilities:

$$P(u|u_t, a_t) \equiv \sum_{\{a_t^d\}}\sum_{\{b_t^d\}} \prod_{d=0}^{D-2} P_b(b_t^d; v_t^d) P_a(a_t^{d+1}; u_t^{d+1}) \cdot P_b(b_t^{D-1}; v_t^{D-1}) \quad (3.22)$$

as

$$Q_a(u_t, a_t) = \sum_{u \in U_D(u_t, a_t)} P(u|u_t, a_t) H_a(u; \omega) \quad (3.23)$$

and

$$V_a(u_t) = \sum_{a_t} P_a(a_t; u_t) \sum_{u \in U_D(u_t, a_t)} P(u|u_t, a_t) H_a(u; \omega). \quad (3.24)$$

$U_D(u_t, a_t)$ is a set of leaf nodes at $d=D$ in the subtree after selecting $a_t$ at $u_t$. $P(u|u_t, a_t)$ is the probability that the node value of leaf node $u$ is backed up to the node value of root node $u_t$. For this reason, $P(u|u_t, a_t)$ is called a *backup probability*. On the other hand, the transition probability where $P_a(a;u)$ and $P_b(b;v)$ are replaced with $\pi_a(a|u)$ and $\pi_b(b|v)$ in $P(u|u_t, a_t)$ is a probability that node $u$ is selected from root $u_t$ through move $a_t$. $P(u|u_t, a_t)$ is called a *realization probability* [n]. The realization probabilities control the search depth in Gekisashi [5][11].

### 3.5 MCSS in multiplayer games

In Section 3.4, we applied MCSS to two-player games like shogi. This section deals with *n*-player games with perfect information. Assume that player 1 plays with players $i$ (=2,3,…,*n*) who take turns in ascending order of $i$. $v(i)$ is the state of player $i$'s turn, and $b(i)$ is its move. If the backup policy of player $i$ is $P_{b(i)}(b(i);v(i))$, then the move and node values are defined in such recursive expressions as Eqs. (3.16) and (3.20) as follows:

$$Q_a(u_t^d, a_t^d) = \sum_{b_t^d(2),\dots,b_t^d(n)} \prod_{i=2}^n P_{b(i)}\left(b_t^d(i); v_t^d(i)\right) \cdot \sum_{a_t^{d+1}} P_a(a_t^{d+1}; u_t^{d+1}) Q_a(u_t^{d+1}, a_t^{d+1}). \quad (3.25)$$

$$V_a(u_t^d) = \sum_{a_t^d} P_a(a_t^d; u_t^d)$$

$$\cdot \sum_{b_t^d(2),\dots,b_t^d(n)} \prod_{i=2}^n P_{b(i)}\left(b_t^d(i); v_t^d(i)\right) V_a(u_t^{d+1}). \quad (3.26)$$

Only defining the move and node values of the leaves by Eqs. (3.17) and (3.21) is necessary as in a two-player game. Therefore, the theories in Sections 3 and 4 are easily applied to *n*-player games.

## 4. Learning Positional Evaluation Functions Based on MCSS

Many methods have been proposed for the learning evaluation functions of computer chess, go, and shogi, based on supervised learning, reinforcement learning, regression learning, and search bootstrapping. In this section, we use backup policy functions $P_a(a;u)$, node values $V_a(s)$, and move values $Q_a(s,a)$ in these learning algorithms.

### 4.1 Supervised learning

The *Bonanza method*, a well-known supervised learning algorithm for positional evaluation functions in shogi, uses game records between professional shogi players as teaching data [12][13]. As an extension of this method, a supervised learning method using policy gradients was proposed by Konemura et al. [14].

In Ref. 14, a loss function to be minimized was defined by

$$\delta_{KLD}(S; \pi^*, \pi) \equiv \sum_{s \in S} \sum_{a \in A(s)} \pi^*(a|s) \ln\left\{\frac{\pi^*(a|s)}{\pi(a|s; \omega)}\right\}. \quad (4.1)$$

$\pi^*$ and $\pi$ are the probability distribution functions of the teacher and learning systems. The Kullback-Leibler divergence is used to measure the difference between the two distribution functions in Eqs. (4.1). $S$ is a set of training positions and $A(s)$ is a set of legal moves at position $s$. $\omega$ denotes a parameter vector included in a positional evaluation function.

Policy $\pi$, which determines the actions of the learning system, is given by a Boltzmann distribution function in Ref. 14 as follows:

$$\pi(a|s; \omega) = exp(H_a(s^*|a, s; \omega)/T)/Z \quad (4.2)$$

and

$$Z \equiv \sum_{x \in A(s)} exp(H_a(s^*|x, s; \omega)/T). \quad (4.3)$$

Position $s^*$ is the leaf node of the PV after move $a$ is selected at position $s$. Its static value is given by $H_a(s^*|a, s; \omega)$. [o]

In this paper, we use backup policy $P_a(a_t;u_t)$ in Eq. (3.5) as

---

n) The realization probability defined in Gekisashi is represented by the features of moves instead of the positional evaluation functions and used as a node-selection policy. However, MCSS computes a node-selection policy only from positional evaluation functions [4].

o) Quiescence search was used in Ref. 14 instead of searching for PVs to reduce the computation time.

policy $\pi$ in Eq. (4.1) considering important variations other than PVs. Therefore, we define the loss function:

$$\delta_{KLD}(S; \pi^*, P_a) \equiv \sum_{s \in S} \sum_{a \in A(s)} \pi^*(a|s) \ln\left\{\frac{\pi^*(a|s)}{P_a(a;s)}\right\}, \quad (4.4)$$

which leads to the following learning rule:

$$\Delta \omega = -\varepsilon \nabla_\omega \delta_{KLD}(S; \pi^*, P_a), \quad (4.5)$$

$$= \varepsilon \sum_{s \in S} \sum_{a \in A(s)} \pi^*(a|s) \nabla_\omega \ln P_a(a;s), \quad (4.6)$$

$$= \frac{\varepsilon}{T_a} \sum_{s \in S} \sum_{a \in A(s)} \{\pi^*(a|s) - P_a(a;s)\} \nabla_\omega Q_a(s,a). \quad (4.7)$$

The calculation of gradient vector $\nabla_\omega Q_a(s,a)$ is described in Section 5.

**4.2 Value-based reinforcement learning**

*TD(λ)* is a typical method using value-based reinforcement learning algorithms [15]. We replace the state value function used in the TD(λ) method with node value $V_a(u_t)$, defined by Eq. (3.20), to derive the following learning rule:

$$\Delta \omega = \varepsilon \sum_{t=1}^{L_a - 1} e^\lambda(t) \delta_t, \quad (4.8)$$

$$e^\lambda(t) = \gamma \lambda e^\lambda(t-1) + \nabla_\omega V_a(u_t), \quad (4.9)$$

$$\delta_t \equiv r_{t+1} + \gamma V_a(u_{t+1}) - V_a(u_t). \quad (4.10)$$

$t$ is the $t$'th time step when the learning agent takes its turn. $L_a$ is the number of steps until the end of the game. $\gamma$ and $\lambda$ are the constants between 0 and 1. $r_t$ is a reward signal, such as win or lose information digitalized to 0 or 1. Computing method $\nabla_\omega V_a(u_t)$ will be described in Section 5.

*Q-learning*, which is another typical method for value-based reinforcement learning algorithms, uses an action value function instead of a state value function. We propose a softmax operation as a backup policy in Q-learning. That is, we use move value function $Q_a(s,a)$, defined in Eq. (3.16), instead of the usual Q-function.

First, we define a loss function of learning:

$$\delta_Q(S,A)$$

$$\equiv \frac{1}{2} \sum_{s \in S, a \in A} \left[ Q_a(s,a) - \sum_b P_b(b;v) \sum_{a'} P_a(a';s') Q_a(s',a') \right]^2. \quad (4.11)$$

$v$ is the state of the opponent's turn after the learning agent makes move $a$ at $s$. $s'$ is the next state after the opponent makes $b$ at $v$. The second term in the square bracket in Eq. (4.11) becomes $V_b(v)$ using Eqs. (3.9) to (3.12). We also use a steepest descent method to derive a learning rule here. The second term is the target value of $Q_a(s,a)$ and should *not* be differentiated with respect to $\omega$.

The learning rule is expressed as

$$\Delta \omega = -\varepsilon \nabla_\omega \delta_Q(S,A) \quad (4.13)$$

$$= -\varepsilon \sum_{s \in S, a \in A} [Q_a(s,a) - V_b(v)] \nabla_\omega Q_a(s,a). \quad (4.14)$$

The calculation of gradient vector $\nabla_\omega Q_a(s,a)$ is described in Section 5. Note that $s$ is not restricted to an actual position that appears during games, but it is an internal node of a search tree.

**4.3 Policy-based reinforcement learning**

*REINFORCE* by Williams is a method of policy-based reinforcement learning [16]. The *PG expectation* method is a learning scheme based on REINFORCE and softmax search. Policy Gradient with Principal Leaf *(PGLeaf)* is a fast approximate version of the PG expectation method [17][18].

Using backup policy $P_a(a_t;u_t)$ and move value $Q_a(u,a)$ in Eq. (3.5), the learning rule in the PG expectation method is expressed:

$$\Delta \omega = \varepsilon \cdot r \sum_{t=0}^{L_a - 1} e_\omega(t) \quad (4.15)$$

$$e_\omega(t) \equiv \nabla_\omega \ln P_a(a_t; u_t)$$

$$= \left[ \nabla_\omega Q_a(u_t, a_t) - \sum_{a \in A(u_t)} P_a(a; u_t) \nabla_\omega Q_a(u_t, a) \right] / T_a. \quad (4.16)$$

The calculation of $\nabla_\omega Q_a(s,a)$ on the right-hand side is described in Section 5.

**4.4 Regression learning**

Let $r(\sigma)$ be a variable that takes 1 if a learning agent wins game $\sigma$, and otherwise it is 0. Next we define function $P_{win}(s)$ that predicts winning the game at state $s$ using node value function $V_a(s)$:

$$P_{win}(s) = 1/[1 + e^{-V_a(s)/\tau}]. \quad (4.17)$$

If a loss function is defined as

$$\delta_{reg}(\sigma) \equiv \frac{1}{2} \sum_{t=1}^{L_a - 1} [r(\sigma) - P_{win}(u_t)]^2, \quad (4.18)$$

then the steepest descent method gives the following learning rule:

$$\Delta \omega = \varepsilon \sum_{t=1}^{L_a-1} [r(\sigma) - P_{win}(u_t)] \nabla_\omega P_{win}(u_t), \quad (4.19)$$

where

$$\nabla_\omega P_{win}(u_t) = \\ (1/\tau)[1 - P_{win}(u_t)] P_{win}(u_t) \nabla_\omega V_a(u_t). \quad (4.20)$$

Alpha Zero [2] also uses a similar regression learning. However, it used a positional evaluation function $H_a(u_t; \omega)$ instead of $V_a(u_t)$ in Eq. (4.17).

### 4.5 Learning by bootstrapping

Veness et al. proposed for chess a learning algorithm called *Search Bootstrapping* [19] and used deep search results to learn evaluation function $H_a(u_t; \omega)$. We apply node values $V_a(s)$ to their bootstrapping approach and define a loss function on set of states $S$:

$$\delta_{bts}(S) \equiv \tfrac{1}{2} \sum_{s \in S}[H_a(s; \omega) - V_a(s)]^2. \quad (4.21)$$

The steepest descent method gives the following learning rule if $V_a(s)$ is the target values of $H_a(s; \omega)$:

$$\Delta \omega = -\varepsilon \sum_{s \in S} [H_a(s; \omega) - V_a(s)] \nabla_\omega H_a(s; \omega). \quad (4.22)$$

In Eq. (4.22), $s$ represents an internal node of a search tree. The value of leaf $s^*$ of a PV, $H_a(s^*)$, was used in the past bootstrapping algorithm as the target value of $H_a(s; \omega)$ in Eq. (4.21) instead of $V_a(s)$ [p].

Moreover, we consider a bootstrapping method that matches the distribution of the children's values of $s$ with the deep search results. $P_H(a; s, \omega)$ and $P_a(a; s)$ are distribution functions used for the backup operation based on $H_a(s; \omega)$ and $V_a(s)$, respectively. We define the following loss function to make $P_H(a; s, \omega)$ close to $P_a(a; s)$:

$$\delta_{PP}(S; P_a, P_H) \equiv \sum_{s \in S}\sum_{a \in A(s)} P_a(a; s) \ln\left\{\frac{P_a(a; s)}{P_H(a; s, \omega)}\right\}. \quad (4.23)$$

$P_a(a; s)$ is the deep search result and is used as a target value. The learning rule becomes

$$\Delta \omega = -\varepsilon \nabla_\omega \delta_{PP}(S; P_a, P_H) \quad (4.24)$$

$$= \varepsilon \sum_{s \in S}\sum_{a \in A(s)} P_a(a; s) \nabla_\omega \ln P_H(a; s, \omega) \quad (4.25)$$

$$= \frac{\varepsilon}{T_a} \sum_{s \in S}\sum_{a \in A(s)} \{P_a(a; s) - P_H(a; s, \omega)\} \nabla_\omega H_a(v(a; s)), \quad (4.26)$$

where $v(a; s)$ is the state after making move $a$ at state $s$. Eq. (4.25) is the cross entropy between two probability distribution

---

p) This algorithm is respectively called *RootStrap* and *TreeStrap* when node $s$ is the root node and an internal node on the path of a PV.

---

functions and was used in the loss function of Alpha Zero [2].

## 5. Learning by Monte Carlo Sampling with Backup Policies

### 5.1 Gradient vectors of move and node values

We proposed a backup policy in Section 3 and used the move and node values that were calculated with the backup policy in the traditional learning methods in Section 4. The gradient vectors of move value functions $\nabla_\omega Q_a(s, a)$ are required for supervised learning in Section 4.1, Q-learning in Section 4.2, and policy gradient reinforcement learning in Section 4.3. The gradient vectors of node value functions $\nabla_\omega V_a(s)$ are needed for TD($\lambda$) learning in Section 4.2 and regression learning in Section 4.4.

We describe the calculation of these gradient vectors in this section. The calculation of the move values when state $s$ is position $u_t$ that appears in the actual games was derived from Refs. 4 and 17. Monte Carlo sampling from root node $u_t$ of a search tree was proposed in Ref. 4 to calculate the move values. In Ref. 4, $\nabla_\omega Q_a(s, a)$ is expressed as

$$\nabla_\omega Q_a(u_t^d, a_t^d) = \sum_{b_t^d} P_b(b_t^d; v_t^d) \sum_{a_t^{d+1}} P_a(a_t^{d+1}; u_t^{d+1})$$

$$\cdot [\{Q_a(u_t^{d+1}, a_t^{d+1}) - V_a(u_t^{d+1})\}/T_a + 1]$$

$$\cdot \nabla_\omega Q_a(u_t^{d+1}, a_t^{d+1}). \quad (5.1)$$

If $u_t^d$ is a leaf node of the search tree, i.e., $d=D-1$, then

$$\nabla_\omega Q_a(u_t^{D-1}, a_t^{D-1}) = \sum_{b_t^{D-1}} P_b(b_t^{D-1} | v_t^{D-1}) \nabla_\omega H_a(u_t^D; \omega). \quad (5.2)$$

$V_a(u_t^{d+1})$ in Eq. (5.1) is written as in Eq. (3.9) using $Q_a(u_t^{d+1}, a_t^{d+1})$. To derive Eqs. (5.1) and (5.2) in Ref. 4, backup policy function $P_b(b_t^d; v_t^d)$ of the opponent is fixed during the learning, assuming that $\nabla_\omega P_b(b_t^d; v_t^d) = 0$. That leads to a very simple expression of the learning rule.

The method for calculating $\nabla_\omega V_a(s)$ is expressed recursively:

$$\nabla_\omega V_a(u_t^d) = \sum_{a_t^d} P_a(a_t^d; u_t^d) \sum_{b_t^d} P_b(b_t^d; v_t^d)$$

$$\cdot [\{Q_a(u_t^d, a_t^d) - V_a(u_t^d)\}/T_a + 1] \nabla_\omega V_a(u_t^{d+1}). \quad (5.3)$$

If $u_t^d$ is a leaf node of the search tree, i.e., $d=D$, then

$$\nabla_\omega V_a(u_t^D) = \nabla_\omega H_a(u_t^D; \omega). \quad (5.4)$$

The derivation of Eq. (5.3) is shown in the appendix. $Q_a(u_t^d, a_t^d)$

can be expressed by $V_a(u_t^{d+1})$ using Eq. (3.15).

Equations (5.1) and (5.3) imply that $\nabla_\omega Q_a(s,a)$ and $\nabla_\omega V_a(s)$ are given recursively by the product of the transition probabilities and the factors in the brackets. This means that these gradient vectors were calculated by Monte Carlo sampling with the backup policies on a search tree where the move and node values are given.

Equations (5.2) and (5.4) indicate that parameters $\omega$ included in $H_a(u_t^D; \omega)$ at leaf nodes $u_t^D$ are updated by learning. The leaf node of a PV and its important variations are frequently visited by Monte Carlo sampling. Therefore, parameters $\omega$, which are highly related to those nodes, are updated frequently.

### 5.2 Learning rules expressed by gradient vectors of node value functions

In the last section, the learning rules were expressed by $\nabla_\omega Q_a(s,a)$ and $\nabla_\omega V_a(s)$. The space dimension of the latter vector is much smaller than that of the former. If the former is written by the latter, much memory space can be saved. We rewrote all the learning rules in Section 4 just using $\nabla_\omega V_a(s)$.

We consider supervised learning in Eq. (4.7). Using (A.1), the updating rule in Eq. (4.7) is rewritten:

$$\Delta \omega = \frac{\varepsilon}{T_a} \sum_{a \in A(u)} \{\pi^*(a|u) - P_a(a;u)\}$$
$$\cdot \sum_b P_b(b; v(a;u)) \nabla_\omega V_a(u(b;v)), \quad (5.5)$$

where $v(a;u)$ is the state after move $a$ at state $u$ and $u(b;v)$ is the state after move $b$ at state $v$.

Next, updating rules (4.15) and (4.16) of the policy gradient reinforcement learning are rewritten:

$$\Delta \omega = \varepsilon \cdot r \sum_{t=0}^{L_a-1} e_\omega(t) \quad (5.6)$$

$$e_\omega(t) = \frac{1}{T_a} \Bigg[ \sum_{b_t} P_b(b_t; v_t(a_t; u_t)) \nabla_\omega V_a\left(u_t^1(b_t; v_t(a_t; u_t))\right)$$
$$- \sum_{a \in A(u_t)} P_a(a; u_t) \sum_{b_t} P_b(b_t; v_t(a; u_t)) \nabla_\omega V_a\left(u_t^1(b_t; v_t(a; u_t))\right) \Bigg]. \quad (5.7)$$

The learning rules of the TD($\lambda$) method and regression learning are represented only by $\nabla_\omega V_a(s)$, as shown in Eqs. (4.9) and (4.20).

Q-learning in Section 4.2 is reduced to a bootstrapping method using states where the opponent selects a move for the following reason. $Q_a(s,a)$ is the value of move $a$ at state $s$ and should be approximated by $H_a(v(a;s); \omega)$ as precisely as possible. For this purpose, we replace $Q_a(s,a)$ with $H_a(v(a;s); \omega)$ in Eq. (4.14):

$$\Delta \omega = -\varepsilon \sum_{s \in S, a \in A} [H_a(v(a;s); \omega) - V_b(v)]$$
$$\cdot \nabla_\omega H_a(v(a;s); \omega). \quad (5.8)$$

This updating rule is reduced to the bootstrapping rule when $s$ is replaced with opponent's state $v$ in Eq. (4.22).

For bootstrapping, the gradient vectors of the node and move values are not necessary for the learning rules in Eqs. (4.22) and (4.26). Note that bootstrapping methods can learn not only the positions that appear in real games but also the internal nodes visited by the Monte Carlo sampling proposed in this section. This is a huge advantage to reduce the number of games necessary for sufficient learning.

### 5.3 Combination of learning methods: including supervised learning

We expressed the updating rules of the major learning methods in shogi using $\nabla_\omega V_a(s)$ and carried them out by Monte Carlo sampling with backup policies on the search tree. The next problem is how to combine these learning methods. In this paper, we divide that problem into two cases depending on whether supervised learning was included or not.

In the former case, we propose the following combination of supervised learning methods using policy gradients [14], Q-learning, and bootstrapping:

Example of learning with a teacher signal:

1) Prepare training data that consist of a set of states $S$ and target distribution $\pi^*(a|s)$ for supervised learning.
2) Do the following procedure for $\forall s \in S$:
{
  2-1) Search for the game tree by MC Softmax search from root node $s$ and preserve the node values at all the nodes visited during the search.
  2-2) Repeat Monte Carlo sampling with backup policies $P_a(a;s)$ and $P_b(b;v)$ from the state after making move $a$ at $s$. During the sampling, do the following procedures:
  {
  - Compute $\Delta \omega$ in Eqs. (4.22) and (5.8) at each node on the sampled path for bootstrapping and Q-learning.
  - Compute $\Delta \omega$ in Eq. (5.5) using Eqs. (5.3) and (5.4) at the leaf node of the sampled path for supervised learning.
  }
  2-3) $\omega$ is updated to $\omega + \Delta \omega$.
}

### 5.4 Combination of learning methods: without teacher signals

If there is no teacher signal on the states and the moves, we combine reinforcement and regression learning based on the game's results with Q-learning and bootstrapping as follows:

【Example of learning without teacher signals】

1) Play a game using MC Softmax Search. Preserve the search trees whose root nodes are piece positions $\{u_t\}$ that appeared in the game and the values of all the nodes included in the search trees.
2) Conduct Monte Carlo sampling from each $u_t$ with backup policies $P_a(a;s)$ and $P_b(b;v)$. During the sampling, do the following procedures:
{
- Compute $\Delta\omega$ in Eqs. (4.22) and (5.8) at each node on the sampled path for bootstrapping and Q-learning.
- Compute $\nabla_\omega V_a(u_t)$ and $\nabla_\omega V_a(u_t^1)$ using Eqs. (5.3) and (5.4) from gradient vector $\nabla_\omega H_a(u^D;\omega)$ at leaf node $u^D$ of the sampled path.
}
3) Compute $\Delta\omega$ respectively using Eqs. (4.8), (5.6), and (4.19) from the results computed in 2) for the TD ($\lambda$)-learning, policy gradient learning, and regression learning. Combine the three values of $\Delta\omega$ by an integrating calculation as a linear summation.
4) Update $\omega$ to $\omega + \Delta\omega$ and return to 1).

## 6. Conclusion

We made two proposals for MCSS. First, we separated a node-selection policy from a backup policy. This separation allows researchers to freely design a node-selection policy based on their searching strategies and to find the exact principal variation given by minimax search. Second, we proposed new sampling-based learning algorithms for a positional evaluation function. This sampling is performed with backup policies in the search tree produced by MCSS. The proposed learning algorithm is applicable to supervised learning, reinforcement learning, regression learning, and search bootstrapping.

The new learning methods consider not only the PV leaves but also the leaves of the important variations around the PVs. The feature parameters highly related to the positions of those leaf nodes can be updated toward their optimal values. The internal nodes in a search tree are learned by bootstrapping. We showed examples of combining these learning methods to reduce the number of games necessary for learning. We are planning to implement our proposed search and learning methods into shogi programs to evaluate their effectiveness. Designing and learning selection strategies in a node-selection policy are also future problems.

## References

[1] Browne, C. B., et al. A Survey of Monte Carlo Tree Search Methods. IEEE Trans. of Comp. Intell. and AI in Games. 2012, vol. 4, no. 1, pp. 1-43.
[2] Silver, D. et al. Mastering Chess and Shogi by Self-Play with a General Reinforcement Learning Algorithm. 2017, arXiv: 1712.01815.
[3] Kocsis, L. and Szepesvari, C. Bandit based Monte-Carlo Planning. In: Furnkranz J., Scheffer T., Spiliopoulou M. (eds) Machine Learning: ECML 2006, pp. 282-293.
[4] Kirii, A., Hara, Y., Igarashi, H., Morioka, Y. and Yamamoto, K. Stochastic Selective Search Applied to Shogi. Proceedings of the 22th Game Programming Workshop (GPW2017), 2017, pp. 26-23. (in Japanese)
[5] Tsuruoka, Y., Yokoyama, D. and Chikayama, T. Game-tree Search Algorithm Based on Realization Probability. ICGA Journal, 2002, pp. 146-153.
[6] Igarashi, H., Morioka, Y. and Yamamoto, K. Learning Positional Evaluation Functions and Simulation Policies by Policy Gradient Algorithm. IPSJ SIG Technical Report. 2013, Vol. 2013-GI-30, No. 6, pp. 1-8. (in Japanese)
    This paper was translated into English and published as Igarashi, H., Morioka, Y. and Yamamoto, K. Reinforcement Learning of Positional Evaluation Functions and Simulation Policies by Policy Gradient Algorithm. The Research Reports of Shibaura Institute of Technology, Natural Sciences and Engineering. 2015, Vol. 58, No.1, pp. 1-9. ISSN 0386-3115.
[7] Auer, P., Cesa-Bianchi, N., and Fisher, P. Finite-time Analysis of the Multiarmed Bandit Problem. Machine Learning. 2002, vol. 47, pp. 235-256.
[8] Sato, Y. and Takahashi, D. A Shogi Program Based on Monte-Carlo Tree Search. IPSJ Journal. 2009, Vol. 50, No. 11, pp. 2740-2751. (in Japanese)
[9] Jaynes, E. T. Information Theory and Statistical Mechanics. Physical Review, 1957, vol.106, no.4, pp.620-630.
[10] Peret, L. and Garcia, F. On-line Search for Solving Markov Decision Process via Heuristic Sampling. In: de Mantaras, R. L., (ed.), ECAI, 2004, pp. 530-534.
[11] Hara, Y., Igarashi, H., Morioka, Y. and Yamamoto, K. A Simple Game-Tree Search Algorithm Based on Softmax Strategy and Search Depth Control Using Realization Probability. Proceedings of the 21th Game Programming Workshop (GPW2016), 2016, pp. 108-111. (in Japanese)
[12] Hoki, K. Source Codes of Bonanza 4.1.3. In: Matsubara, H., (ed.), Advances in Computer Shogi 6, Kyoritsu Shuppan, 2012, Chap. 1, pp. 1-23. (in Japanese)
[13] Hoki, K. and Kaneko, T. Large-Scale Optimization for Evaluation Functions with Minimax Search. Journal of Artificial Intelligence Research. 2014, vol. 49, pp. 527-568.
[14] Konemura, H., Yamamoto, K., Morioka, Y. and Igarashi, H. Policy Gradient Supervised Learning of Positional Evaluation Function in Shogi: Using Quiescence Search and AdaGrad. Proceedings of the 22th Game Programming Workshop (GPW2017), 2017, pp. 1-7. (in Japanese)
[15] Sutton, R. S. and Barto A. G. Reinforcement Learning: An Introduction. The MIT Press, Massachusetts, 1998, 322p.
[16] Williams, R. J. Simple Statistical Gradient- Following Algorithms for Connectionist Reinforcement Learning. Machine Learning,

1992, vol. 8, pp. 229-256.

[17] Igarashi, H., Morioka, Y. and Yamamoto, K. Learning Static Evaluation Functions Based on Policy Gradient Reinforcement Learning. Proceedings of the 17th Game Programming Workshop (GPW2012), 2012, pp. 118-121. (in Japanese)

[18] Morioka, Y. and Igarashi, H. Reinforcement Learning Algorithm that Combines Policy Gradient Method with Alpha-Beta Search. Proceedings of the 17th Game Programming Workshop (GPW2012), 2012, pp. 122-125. (in Japanese)

[19] Veness, J., Silver, D., Uther, W. and Blair, A. Bootstrapping from Game Tree Search. Advances in Neural Information Processing Systems 22, 2009.

## Appendix

### A.1 Recursive expression on gradient vectors of node values

Recursive expression on $\nabla_\omega Q_a(s, a)$ in Eq. (5.1) was given in Ref. 4. This section derives Eq. (5.3) from Eq. (5.1). The left-hand side of the latter is written using Eq. (3.15):

$$\nabla_\omega Q_a(u_t^d, a_t^d) = \sum_{b_t^d} P_b(b_t^d; v_t^d) \cdot \nabla_\omega V_a(u_t^{d+1}). \quad (A.1)$$

If $d$ is replaced with $d+1$ in (A.1),

$$\nabla_\omega Q_a(u_t^{d+1}, a_t^{d+1}) = \sum_{b_t^{d+1}} P_b(b_t^{d+1}; v_t^{d+1}) \cdot \nabla_\omega V_a(u_t^{d+2}). \quad (A.2)$$

Substituting (A.2) into the right-hand side of Eq. (5.1), we get

$$\nabla_\omega Q_a(u_t^d, a_t^d) =$$
$$\sum_{b_t^d} P_b(b_t^d; v_t^d) \sum_{a_t^{d+1}} P_a(a_t^{d+1}; u_t^{d+1})$$
$$\cdot [\{Q_a(u_t^{d+1}, a_t^{d+1}) - V_a(u_t^{d+1})\}/T_a + 1]$$
$$\cdot \sum_{b_t^{d+1}} P_b(b_t^{d+1}; v_t^{d+1}) \cdot \nabla_\omega V_a(u_t^{d+2}). \quad (A.3)$$

Comparing the operand of $\sum_{b_t^d} P_b(b_t^d; v_t^d) \cdot$ in (A.1) with that in (A.3),

$$\nabla_\omega V_a(u_t^{d+1}) = \sum_{a_t^{d+1}} P_a(a_t^{d+1}; u_t^{d+1})$$
$$\cdot [\{Q_a(u_t^{d+1}, a_t^{d+1}) - V_a(u_t^{d+1})\}/T_a + 1]$$
$$\cdot \sum_{b_t^{d+1}} P_b(b_t^{d+1}; v_t^{d+1}) \cdot \nabla_\omega V_a(u_t^{d+2}). \quad (A.4)$$

If $d+1$ is replaced with $d$ in (A.4), EQ. (5.3) is derived.